\documentclass{article}

\usepackage[preprint]{neurips_2025}

\usepackage[utf8]{inputenc} 
\usepackage[T1]{fontenc}    
\usepackage{hyperref}       
\usepackage{url}            
\usepackage{booktabs}       
\usepackage{amsfonts}       
\usepackage{nicefrac}       
\usepackage{microtype}      
\usepackage{xcolor}         
\usepackage{graphicx}       
\usepackage{multirow}
\usepackage{multicol}
\usepackage{array}
\usepackage{amsmath}
\usepackage{siunitx}
\usepackage{enumitem}
\usepackage[caption=false]{subfig}
\usepackage{float}
\usepackage[most]{tcolorbox}

\newcommand\dif{\mathop{}\!\mathrm{d}}

\title{AutoBio: A Simulation and Benchmark for Robotic Automation in Digital Biology Laboratory}

\author{%
  Zhiqian~Lan$^{1,2*}$\quad Yuxuan~Jiang$^{1,2*}$\quad Ruiqi~Wang$^{3}$\quad Xuanbing~Xie$^{1}$\quad Rongkui~Zhang$^{3}$  \\
 \textbf{Yicheng~Zhu$^{4}$\quad Peihang~Li$^{1}$\quad Tianshuo~Yang$^{1}$\quad Tianxing~Chen$^{1}$\quad Haoyu~Gao$^{3}$}\\
 \textbf{Xiaokang~Yang$^{4}$\quad Xuelong~Li$^{2}$\quad Hongyuan~Zhang$^{2}$\quad Yao~Mu$^{4}$\quad Ping~Luo$^{1,5\dagger}$}\\
  $^{1}$HKU\quad $^{2}$TeleAI\quad $^{3}$THU\quad $^{4}$SJTU\quad $^{5}$HKU Shanghai Intelligent Computing Center\\
}

\begin{document}
\begin{NoHyper}
    \let\thefootnote\relax
    \footnotetext{$^*$Equal contribution.\quad $^\dagger$Correspondence to pluo@cs.hku.hk\quad \href{https://github.com/autobio-bench/AutoBio}{\textcolor{blue}{Repo: github.com/autobio-bench/AutoBio}} }
\end{NoHyper}

\maketitle


\begin{abstract}

Vision-language-action (VLA) models have shown promise as generalist robotic policies by jointly leveraging visual, linguistic, and proprioceptive modalities to generate action trajectories. While recent benchmarks have advanced VLA research in domestic tasks, professional science-oriented domains remain underexplored. We introduce AutoBio, a simulation framework and benchmark designed to evaluate robotic automation in biology laboratory environments—an application domain that combines structured protocols with demanding precision and multimodal interaction. AutoBio extends existing simulation capabilities through a pipeline for digitizing real-world laboratory instruments, specialized physics plugins for mechanisms ubiquitous in laboratory workflows, and a rendering stack that support dynamic instrument interfaces and transparent materials through physically based rendering. Our benchmark comprises biologically grounded tasks spanning three difficulty levels, enabling standardized evaluation of language-guided robotic manipulation in experimental protocols. We provide infrastructure for demonstration generation and seamless integration with VLA models. Baseline evaluations with two SOTA VLA models reveal significant gaps in precision manipulation, visual reasoning, and instruction following in scientific workflows. By releasing AutoBio, we aim to catalyze research on generalist robotic systems for complex, high-precision, and multimodal professional environments. The simulator and benchmark are \href{https://github.com/autobio-bench/AutoBio}{{publicly available}} 
to facilitate reproducible research.

\end{abstract}

\section{Introduction}

Vision-language-action (VLA) model architectures jointly leverage vision, language, and proprioception modalities to generate action trajectories in an end-to-end manner. By aligning capabilities more closely with how humans perceive and interact with the world, these models hold promise as a pathway toward generalist robotic policies. Recent VLA models \cite{brohan2023rt, kim2024openvla, team2024octo, liu2025rdtb, black2024pi_0, intelligence2025pi_} have demonstrated impressive results in real-world scenarios, including table bussing, clothing folding, and household manipulation. However, current benchmarks remain largely confined to domestic settings, leaving a critical gap in evaluating VLA models for professional, science-oriented scenarios.

Biology laboratories represent a uniquely promising yet challenging environment for robotic automation. Experiments in these environments follow clear, rigorous protocols \cite{schilling2008standardizing}, making them well-suited for language-guided robots to interpret and execute. Additionally, the repetitive and time-consuming nature of many laboratory tasks presents opportunities for automation to reduce researcher workload and enhance efficiency. Existing solutions, such as automated sample preparation platforms \cite{may2016automated}, often prioritize throughput at the expense of flexibility, whereas robotic systems could achieve human-level adaptability. Nevertheless, biological experiments impose distinct challenges on robotic intelligence: Long-horizon workflows require perception and interaction with diverse interfaces (e.g., digital displays, control panels, and articulated mechanisms). Precision-dependent tasks like slot alignment are ubiquitous, while transparent liquids and containers complicate visual reasoning. These characteristics make biology laboratories an ideal benchmark for evaluating VLA capabilities in language grounding, visual understanding, and high-precision manipulation.

To realize this potential, however, it is critical to first address the challenges of simulating biology laboratories—a task that demands advanced asset modeling, physics simulation, and rendering capabilities. Robotic simulations typically use physics engines such as MuJoCo \cite{todorov2012mujoco}, Bullet \cite{coumans2021}, and PhysX to provide a foundation for physics-based interaction. Building upon these engines, each simulator offers its own set of features tailored to specific focus areas. For instance, robosuite \cite{robosuite2020} and MuJoCo Playground \cite{zakka2025mujocoplayground} primarily target reinforcement learning (RL) \cite{sutton1998reinforcement} algorithms through standardized tasks involving pick-and-place operations, articulated object manipulation, and locomotion. RoboTwin \cite{mu2025robotwin} emphasizes dual-arm coordination and proposes an automated pipeline for data synthesis. These simulations generally address out-of-the-box contact-based rigid body interactions, and lack specialized capabilities required for biological experimentation.

\begin{figure}[t]
    \centering
    \includegraphics[width=\linewidth]{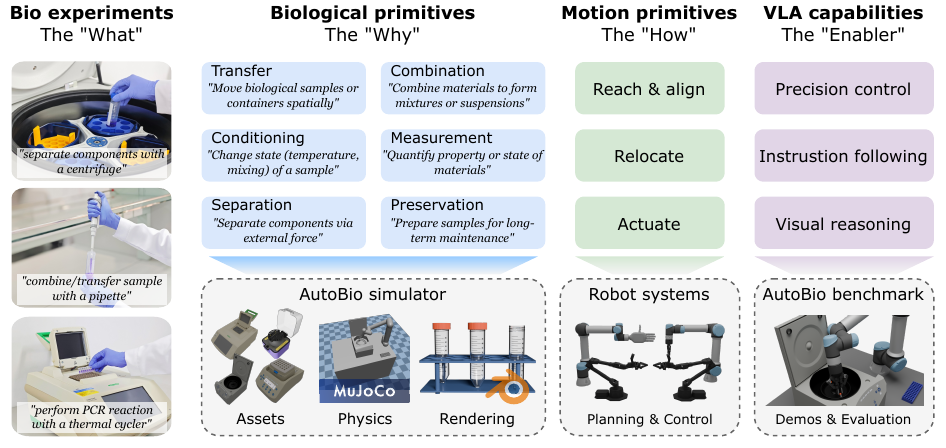}
    \vspace{-2em}
    \caption{
    AutoBio framework. AutoBio decomposes complex experiments into fundamental biological primitives. These are then implemented via robotic motion primitives within a specialized simulation environment. AutoBio simulator features instrument digitization pipeline, custom physics plugins for lab mechanisms, and rendering stack supporting dynamic interfaces and transparent materials. This enables creation of biologically grounded benchmark tasks to evaluate VLA models on precision control, instruction following, and visual reasoning capabilities in scientific workflows.
    }
    \vspace{-1em}
    \label{fig:framework}
\end{figure}

In the AutoBio simulator, we extend asset modeling, physics simulation, and rendering capabilities to streamline the entire pipeline of biological experiment simulation, specifically targeting the biological primitives outlined in Figure \ref{fig:framework}. In biological experiments, many operations rely on specialized instruments such as centrifuges, thermal cyclers, and mixers. We develop a workflow to transform real-world instruments into manipulable assets within AutoBio through 3D Gaussian Splatting \cite{kerbl20233d}, CAD refinement, and texture baking. These transformed instruments are then paired with self-contained logic to faithfully replicate their real-world characteristics. Regarding physics simulation, while the physical world operates under universal laws, physics engines typically provide only generalized implementations, leaving finer-grained interactions to be extended by users. To address this, we develop a suite of MuJoCo plugins supporting interactions prevalent in biological experiments—including thread mechanisms, detent mechanisms, eccentric mechanisms, and quasi-static liquid computation—that are rarely addressed in existing simulators. 
Finally, rendering fidelity proves crucial as vision serves as the primary input modality for VLA models. The prevalence of transparent materials in laboratory environments poses particular challenges, as conventional blend-mode rasterization engines handle transparency poorly. We integrate our simulation with Blender's Physically Based Rendering (PBR) pipeline to achieve visually accurate transparency effects for containers and liquids. 
In addition, we implement dynamic texture rendering for instrument displays, enabling interactive UI manipulation—a critical feature for tasks involving digital interfaces.

By combining the above-mentioned simulation features, we propose the AutoBio benchmark, which distills robotic manipulation primitives into practical biological experiment tasks. We provide comprehensive infrastructure to facilitate trajectory synthesis, demonstration generation, and standardized data interfaces for VLA model integration. Our benchmark comprises tasks across three difficulty levels, where we evaluate two open-source SOTA VLA models ($\pi_0$ \cite{black2024pi_0} and RDT \cite{liu2025rdtb}). This evaluation reveals critical limitations in current approaches and suggests potential improvements in model architecture and training methodologies.

Our key contributions summarize as follows:
(1) A simulator designed for biology labs, featuring instrument digitization, physics plugins (thread/detent mechanisms, quasi-static liquids), and PBR rendering for transparency and reactive displays.
(2) A benchmark with biologically grounded tasks, enabling standardized evaluation of robotic automation in lab protocols, with support for trajectory synthesis and VLA integration.
(3) Systematic evaluation of VLA models in science-oriented settings, revealing critical gaps in precision manipulation, instruction following and visual reasoning.

\begin{table}[t]
\centering
\caption{Comparison of AutoBio with related robotic manipulation benchmarks}
\label{tab:compare}
\footnotesize
\setlength{\tabcolsep}{3pt}
\begin{tabular}{@{}lcccccccccc@{}}
\toprule
{\scriptsize \begin{tabular}[c]{@{}c@{}}Benchmark\\ (Simulator)\end{tabular}  }  &
{\scriptsize \begin{tabular}[c]{@{}c@{}}Target \\ domain\end{tabular}  } &
{\scriptsize \#Tasks  } & 
{\scriptsize \begin{tabular}[c]{@{}c@{}}Interactive\\ instrument\end{tabular}  } &
{\scriptsize \begin{tabular}[c]{@{}c@{}}Threaded \\ object\end{tabular}  } &
{\scriptsize  Fluid  } & 
{\scriptsize \begin{tabular}[c]{@{}c@{}}Reactive\\ display\end{tabular}  } &
{\scriptsize \begin{tabular}[c]{@{}c@{}}Render\\ backend\end{tabular}  } &
{\scriptsize \begin{tabular}[c]{@{}c@{}}Demo\\ synthesis\end{tabular}  } &
{\scriptsize \begin{tabular}[c]{@{}c@{}}VLA model\\ train \& eval\end{tabular}  }
\\ \midrule

Meta-world \cite{yu2019meta}                                                                         & -                                                        & 50      & $\times$                                                              & $\times$                                                         & $\times$   & $\times$                                                        & MuJoCo                                                      & \checkmark                                                      & $\times$                                                                  \\
Robosuite  \cite{robosuite2020}                                                                         & -                                                        & 9       & $\times$                                                              & $\times$                                                         & $\times$   & $\times$                                                        & Isaac Sim                                                   & $\times$                                                      & $\times$                                                                  \\
Factory  \cite{NarangSAMRWGMSL2022factory}                                                                           & Factory                                                  & 8       & $\times$                                                              & \checkmark                                                         & $\times$   & $\times$                                                        & Isaac Gym                                                   & $\times$                                                      & $\times$                                                                  \\
Maniskill 2  \cite{gu2023maniskill2}                                                                       & -                                                        & 20      & $\times$                                                              & $\times$                                                         & $\times$   & $\times$                                                        & SAPIEN                                                      & \checkmark                                                      & $\times$                                                                  \\
Robotwin   \cite{mu2025robotwin}                                                                         & -                                                        & 14      & $\times$                                                              & $\times$                                                         & $\times$   & $\times$                                                        & SAPIEN                                                      & \checkmark                                                      & \checkmark                                                                  \\
Chemistry3D   \cite{li2024chemistry3d}                                                                      & Chemistry                                                & 5       & $\times$                                                              & $\times$                                                         & \checkmark   & $\times$                                                        & Isaac Sim                                                   & $\times$                                                      & $\times$                                                                  \\
AutoBio (ours)                                                                      & Biology                                                  & 16      & \checkmark                                                              & \checkmark                                                         & \checkmark   & \checkmark                                                        & Blender                                                     & \checkmark                                                      & \checkmark                                                         \\ \bottomrule
\end{tabular}
\end{table}

\section{Related Works}
\paragraph{Vision-Language-Action Model.}
Recent generalist policies integrate vision, language, and control for broad task generalization. RT-2 \cite{brohan2023rt} leverages web-scale vision-language data for robotic control, while OpenVLA \cite{kim2024openvla}—trained on 1M real-world demos—surpasses larger closed models on various manipulation tasks. RDT \cite{liu2025rdtb} extends diffusion transformers to bimanual manipulation, enabling zero-shot generalization to novel objects and scenes. $\pi_0$ \cite{black2024pi_0} (and its successor $\pi_{0.5}$ \cite{intelligence2025pi_}) combines VLMs with flow matching for smooth, cross-embodiment action generation. However, the benchmark tasks typically involve coarse manipulation (picking, placing, folding) rather than precise lab procedures. Long-horizon, high-precision tasks and interacting with digital interfaces are rarely evaluated. AutoBio is intended to fill this gap by focusing on lab-specific objects and instruments, explicitly testing fine-grained vision-language-action reasoning not covered by prior work.

\paragraph{Robotic Manipulation Benchmark.}
Existing benchmarks like ManiSkill \cite{gu2023maniskill2}, Meta-World \cite{yu2019meta}, robosuite \cite{robosuite2020}, ARMBench \cite{mitash2023armbench}, and Factory \cite{NarangSAMRWGMSL2022factory} focus on rigid-object manipulation (e.g., pick/place, assembly) in home, warehouse or factory settings. While Chemistry3D \cite{li2024chemistry3d} introduces science-oriented tasks for chemical experiments, most lack support for fluids, digital interfaces, or lab-specific precision. AutoBio addresses this gap by simulating biological workflows with fluid handling, transparent materials, and interactive instrument UIs—challenges absent in prior benchmarks. A feature comparison with related benchmarks can be found in Table \ref{tab:compare}.

\paragraph{Robotic Automation in Laboratory.}
A parallel research thread builds specialized ``self-driving'' labs for science. Szymanski et al.~\cite{szymanski2023autonomous} describe A‑Lab that fully automates solid-state materials synthesis with fixed hardware pipelines tailored to inorganic chemistry. In their system, robotic arms dose powders, handle furnaces, and transfer samples between instruments while ML algorithms propose recipes. Similarly, other efforts have automated enzyme engineering or pharmaceutical screens \cite{holland2020automation, rapp2024self, tom2024self}. However, these platforms typically prioritize throughput: each piece of equipment is specialized for a specific protocol, and human oversight is still needed for setup or maintenance \cite{holland2020automation}. AutoBio targets the complementary goal of flexible, human-like manipulation in the lab. Instead of crafting one-purpose machines, AutoBio’s framework is designed to move toward handling novel protocols end-to-end: reading digital instructions, applying proper tools, and adapting on the fly, that lie beyond the scope of existing lab automation work.

\section{AutoBio Simulator}
\subsection{AutoBio Assets}
\begin{figure}[t]
    \centering
    \includegraphics[width=0.95\linewidth]{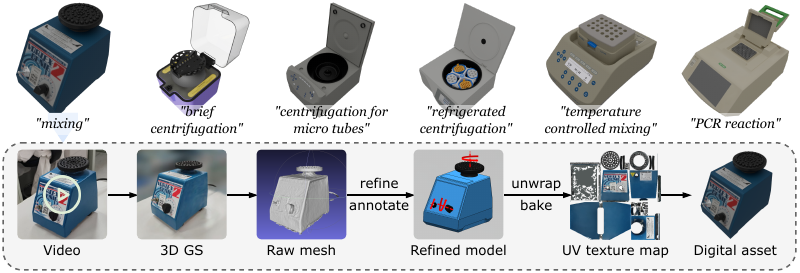}
    \caption{Digitized instruments for fundamental biological experiment operations, with an example taken from vortex mixer demonstrating the proposed workflow for digitizing real-world instruments.}
    \label{fig:asset}
\end{figure}
With reference to a real-world biomedical laboratory, AutoBio constructs dimensionally accurate digital models of common biological experimental assets and integrates them into a virtual environment. To support robotic manipulation, AutoBio meticulously preserves the physical interaction properties of these digital assets—including collision characteristics and articulated relationships—to ensure physical fidelity in simulation. All digital assets, along with their associated physical and visual properties, are formally described using the MJCF modeling language.

\begin{table}[h]
    \caption{Categories of AutoBio assets}
    \label{tab:assets}
    \centering
    \begin{tabular}{@{}lll@{}}
    \toprule
    Type     & Function                                 & Asset                                             \\ \midrule
    Instrument & Perform experimental procedures          & Centrifuge,  Thermal Cycler, Mixer, Pipette, \ldots \\
    Container     & Handle and store samples              & Centrifuge tube, Cryovial, Pipette tip, \ldots                         \\
    Rack           & Host containers and tools                & Multiple-slot rack, Pipette rack, Tip box              \\
    Robot          & Execute manipulation                     & UR5e, Aloha, Robotiq gripper, DexHand                                   \\
    \bottomrule
    \end{tabular}
\end{table}

Our assets are broadly categorized into four classes, as listed in Table~\ref{tab:assets}. To ensure the visual and geometrical fidelity of the instrument models, we propose a pipeline that incorporates 3D Gaussian Splatting (3DGS) \cite{kerbl20233d} for modeling laboratory assets. As illustrated in Figure \ref{fig:asset}, the process begins with multi-view video capture of real-world instruments. We then apply the PGSR algorithm \cite{chen2024pgsr} to reconstruct high-quality 3DGS assets. Coarse 3D meshes are subsequently extracted to achieve surface reconstruction.
The raw mesh from 3DGS often contains redundant vertices and irregular topology due to the discrete nature of Gaussian splats. To optimize the mesh for physical simulation, we refine it in CAD modeling software, producing a simplified, watertight low-poly version while preserving critical geometric features. Articulations are also annotated during this stage through mating constraints. The simplified model then undergoes an automated pipeline that UV-unwraps the mesh, bakes vertex colors from the original high-poly mesh onto a UV texture map, and organizes the output as a digital asset compatible with MuJoCo.

\subsection{AutoBio Physics}

AutoBio employs MuJoCo as its physics engine to simulate rigid-body dynamics, including interactions among robots, labware and instruments. This framework supports accurate multi-articulated system modeling with rich contact, which are essential for replicating laboratory operations such as pipetting, tube handling, and instrument manipulation. In addition to MuJoCo’s native physics features, AutoBio incorporates customized plugins to efficiently simulate laboratory-specific physical behaviors. As depicted in Figure \ref{fig:autobio physics}, these plugins include:
\begin{figure}
    \centering
    \subfloat[Thread mechanism]{\label{fig:thread}\includegraphics[width=0.23\linewidth]{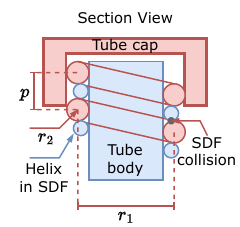}}%
    \hspace{0.01\linewidth}%
    \subfloat[Detent mechanism]{\label{fig:detent}\includegraphics[width=0.23\linewidth]{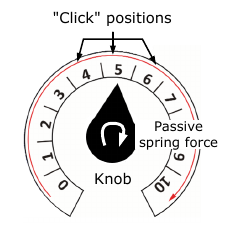}}%
    \hspace{0.01\linewidth}%
    \subfloat[Eccentric mechanism]{\label{fig:eccentric}\includegraphics[width=0.23\linewidth]{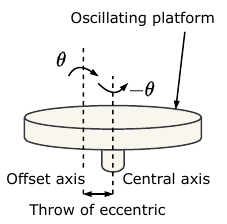}}%
    \hspace{0.01\linewidth}%
    \subfloat[Quasi-static liquid]{\label{fig:liquid}\includegraphics[width=0.23\linewidth]{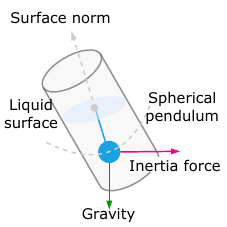}}%
    \caption{AutoBio physics plugins}
    \label{fig:autobio physics}
\end{figure}

\textbf{Thread mechanism} models the assembly and mechanical properties of mating threads (e.g., the cap-body assemblies of centrifuge tubes). Inspired by the separation design of collision and visual geometry, we propose the use of \emph{circular helix's} signed distance function (SDF) to substitute meshes in the collision computation of threads. A parameterized circular helix in space can be expressed as
\begin{equation*}
    H(t; r_1, p)=[x,y,z]^\top=[r_1\cos t, r_1\sin t, pt]^\top.
\end{equation*}
An SDF calculates the orthogonal distance of a given point $P\in\mathbb{R}^3$ to the boundary of a geometric shape, which is an implicit representation for 3D geometry. For an unbounded helix, we provide an approximation to its SDF since there is no analytical solution:
\begin{gather*}
    t_{0}=\operatorname{atan2}(P_y, P_x),\quad k=\left\lfloor{{(P_z-t_{0}p)}/{(2\pi{}p)}}\right\rceil
    , \quad
    SDF(P) = \left\|H\left(2k\pi + t_{0};r_1, p\right)-P\right\|_2
\end{gather*}
We refer to the appendix for the bounded case.
Contact between threads is enabled by assigning a gauge radius $r_2$ for each helix. The resultant SDF of the composite shape is computed by offsetting the helix's SDF by $-r_2$.
Finally, a helical thread can be configured by a tuple $(r_1, r_2, p, l, h)$. The screw motion can be achieved by rigidly attaching mating helical threads to the nut-bolt pair.

\textbf{Detent mechanism} simulates incremental motion with discrete "click" positions in lab instruments, such as stepped knobs or handles. Specifically, the detent mechanism provides tactile feedback via passive spring force generated through relative displacement with the nearest gear position.

\textbf{Eccentric mechanism} generates oscillating motion through off-axis rotation, enabling realistic simulation of mixers (e.g., vortex mixer). The plugin achieves eccentric orbital motion through negatively coupled rotations about two parallel joint axes.

\textbf{Quasi-static liquid} provides an approximate yet efficient simulation of liquid shape deformation within containers, complementing MuJoCo's limitation in fluid modeling. Specifically, this module treats the liquid surface as a planar interface, neglecting wave propagation and pouring effects. This simplification allows us to describe liquid deformation using only two states: the liquid level height and the surface normal vector. The motion of the surface normal is governed by a damped spherical pendulum system in reponse to external container acceleration. We derive the ODE system of the surface normal through analytical mechanics and Euler-Lagrange equation (see appendix for details).
After the normal vector acquired, the liquid body is computed as the intersection of the oriented halfspace and container inner surface. We compute the surface height through volume conservation constraint, thereby generating liquid geometry within the container.

\subsection{AutoBio Rendering}
\begin{figure}
    \centering
    \subfloat[AutoBio rendering backends]{\label{fig:backends}\includegraphics[width=0.49\linewidth]{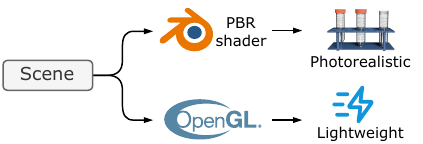}}%
    \hspace{0.05\linewidth}%
    \subfloat[Reactive user interface]{\label{fig:ui}\includegraphics[width=0.35\linewidth]{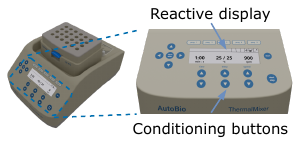}}%
    \caption{AutoBio rendering features}
    \label{fig:rendering}
\end{figure}
AutoBio adopts a flexible rendering strategy, offering two rendering backends to accommodate both fast visualization and photorealistic rendering, as summarized by Figure \ref{fig:backends}. Our simulation features are equally supported by two backends, but with different visual fidelity. Additionally, it implements real-time texture rendering for reactive panel interfaces in laboratory instruments.

\textbf{Basic render} directly utilizes MuJoCo's native OpenGL renderer, which offers fast but limited visualization. Rendering artifacts would occur when visualizing nested transparent objects, such as transparent tubes containing liquids, due to depth sorting errors. So, we fallback to mark some surfaces as opaque for more meaningful results.

\textbf{Advanced render} bridges MuJoCo simulation state to Blender's rendering pipeline, leveraging its physically based rendering (PBR) shaders to achieve photorealistic materials grounded in real-world light behavior. Particularly for container-associated manipulation tasks, this solution enables accurate rendering of transparent materials—including polyethylene, glass, and liquids—through configurable optical parameters such as transmission coefficients, refractive indices, and surface roughness.

\textbf{Reactive user interface} employs dynamically loaded texture maps to render control panels and displays for some lab instruments, providing visual feedback for robotic manipulation, as shown in Figure \ref{fig:ui}. This module maintains full compatibility with both rendering backends described previously.

\section{AutoBio Benchmark}
\begin{figure}[t]
    \centering
    \subfloat[\textbf{Close thermal cycler lid} (easy)]{\label{fig:close}\includegraphics[width=0.95\linewidth]{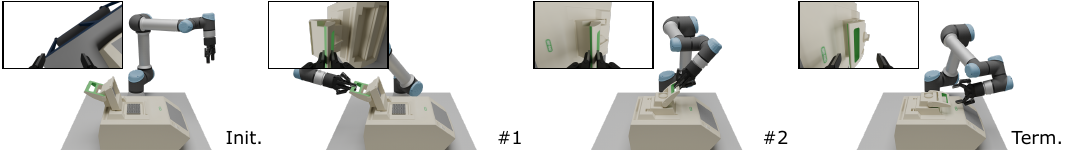}}\\
    \subfloat[\textbf{Unscrew centrifuge tube cap} (medium)]{\label{fig:unscrew}\includegraphics[width=0.95\linewidth]{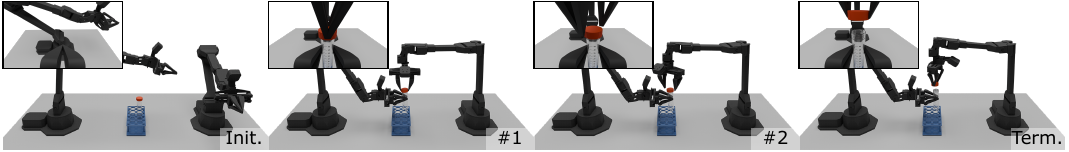}}\\
    \subfloat[\textbf{Operate thermal mixer panel} (hard)]{\label{fig:panel}\includegraphics[width=0.95\linewidth]{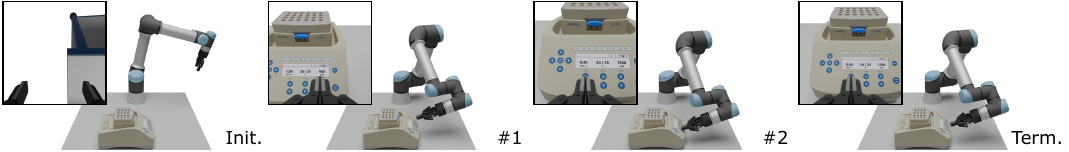}}
    \caption{Task progression across three difficulty levels. Each step includes a bordered inset (top-left) showing supplementary camera perspectives for contextual clarity.}
    \label{fig:benchmark-task}
\end{figure}

Building upon the capabilities of the AutoBio simulator, we develop the AutoBio benchmark, a suite of biologically grounded tasks designed to evaluate robotic automation in laboratory settings. This benchmark focuses on assessing the performance of VLA models in executing common experimental protocols that demand precision, multimodal understanding, and complex manipulation.



To systematically evaluate VLA models, AutoBio consists of 16 tasks that are categorized into three difficulty levels (Figure \ref{fig:benchmark-task}), characterized by progressively increasing demands on precision, language understanding, and visual reasoning:

\begin{itemize}[leftmargin=*]
\setlength\itemsep{0em}
\item \textbf{Level 1 (Easy):} These tasks feature low demands on vision and manipulation precision, exemplified by straightforward actions like closing a thermal cycler lid (\ref{fig:close}). Task requirements and language instructions are generally static, making them suitable for initial system integration and basic policy verification.
\item \textbf{Level 2 (Medium):} Tasks at this level, such as unscrewing a centrifuge tube cap (\ref{fig:unscrew}), present increased challenges in visual perception and manipulation precision. Language instructions may vary based on randomized task parameters (e.g., target position), requiring models to exhibit basic generalization and instruction following capability beyond memorization. These tasks are designed to evaluate fundamental VLA model performance.
\item \textbf{Level 3 (Hard):} High-demand tasks like operating a thermal mixer panel (\ref{fig:panel}) require visual reasoning, fine-grained manipulation, and robust instruction interpretation. Successful completion often necessitates effective closed-loop control and the ability to perform complex cross-modal reasoning. These tasks are intended to rigorously challenge the capabilities of state-of-the-art VLA models in scientifically relevant scenarios.
\end{itemize}

Each benchmark task includes:
\begin{itemize}[leftmargin=*]
\setlength\itemsep{0em}
\item \textbf{Scene Initialization and Randomization:} Logic for setting up the initial environment and introducing variations to task parameters and object configurations, promoting generalization.
\item \textbf{Demonstration Synthesis:} Logic for synthesizing expert trajectories.
\item \textbf{Status Checking:} Mechanisms to determine task success, or progress based on predefined criteria.
\item \textbf{VLA Model Interfacing:} Prompts, along with clearly defined mappings for proprioceptive inputs, action spaces, and camera feeds, to simplify integration with diverse VLA architectures.
\end{itemize}
Task executions, whether from expert demonstrations or learned policies, can be serialized into a compact format. This enables reproducible replays and facilitates offline rendering or analysis, crucial for standardized benchmarking.

The AutoBio benchmark currently supports two primary robotic arm configurations:
\begin{itemize}[leftmargin=*]
\setlength\itemsep{0em}
\item \textbf{Aloha:} This arm is equipped with its native gripper. Characterized by a smaller reach and lower payload capacity, it is primarily employed in tasks involving lighter objects, such as handling individual centrifuge tubes.
\item \textbf{UR5e:} This arm offers greater reach and payload. We provide two end-effector options for the UR5e: (a) \textbf{UR5e-Robotiq:} The UR5e paired with a Robotiq 2F-85 parallel gripper. This configuration is suited for tasks requiring interaction with larger instruments or a wider operational range. (b) \textbf{UR5e-DexHand:} The UR5e equipped with a DexHand 021, a 19-DOF dexterous hand. This setup is intended for tasks demanding more dexterity, such as precise pipette operation. Recognizing that current VLA models are often optimized for simpler, low-DOF end-effectors, we offer a simplified control mode for the DexHand. In this mode, most of its DOFs are pre-configured and fixed, with only the metacarpophalangeal (MCP) joint of the thumb actuated collectively to mimic a gripper-like open/close action.
\end{itemize}
Our motion planning utilities are designed to be robot-agnostic, facilitating the future integration of additional robotic platforms and end-effector configurations. We plan to expand support for full dexterous manipulation, including synthesis and training for dexterous hands, in upcoming releases.

AutoBio benchmark tasks are organized into two main scene layouts: single-arm and dual-arm. In single-arm tasks, the placement of the arm relative to target objects and instruments is flexible, allowing for optimized setups for specific operations. For tasks utilizing two robotic arms, typically positioned at opposite ends of a wide laboratory workbench, mimicking collaborative human workflows. For visual perception, all tasks include multiple camera views. In dual-arm setups, this typically comprises a fixed front-facing global camera and two wrist-mounted cameras (one on each arm) providing egocentric views. Single-arm tasks also use a combination of fixed and wrist-mounted cameras, adaptable to the specific task requirements.

\section{Experiments}

\subsection{Baselines and Experimental Settings}
We choose 3 tasks from each difficulty level (Table \ref{tab:result}) to evaluate VLA capabilities. For each task, we generate 100 demonstration trajectories at a frequency of \qty{50}{Hz} formatted as a LeRobot \cite{cadene2024lerobot} dataset. 
The total training set consists of over 792k frame of data, equivalent to 4.4 hours of continuous recording \footnote{The datasets are published at HuggingFace (\url{https://huggingface.co/autobio-bench}), with two variants for MuJoCo and Blender flavor rendering.}.
All tasks except \textbf{Operate thermal mixer panel} are scored binarily (1 for success, 0 for failure). For \textbf{Operate thermal mixer panel}, we use a relative progress score to better reflect performance difference due to observed policy difficulties. Task details are provided in the appendix.

We evaluate two open-source VLAs: $\pi_0$ \cite{black2024pi_0} and RDT \cite{liu2025rdtb}. We adapt our data to each model's requirements (proprioception dimensions, image history, normalization), then finetune with default fine-tuning configurations starting with their pretrained checkpoints ($\pi_0$-base and RDT-1B). Adaptation details and model differences can be found in the appendix.

To examine data scaling effects, we train each model on both full (100 episodes) and reduced (20 episodes) datasets while maintaining consistent normalization. Each configuration undergoes three seeded runs, trained for 30000 steps with batch size 32. A single run takes between 10 to 14 hours on an NVIDIA H800 GPU, resulting in a total runtime of approximately 1,500 GPU hours.

\subsection{Experimental Results}
\begin{table}[t]
    \newcommand{\SB}{\small\bfseries}
    \addtolength{\tabcolsep}{-0.2em}

\newcommand{\openpipickup}{$13.3 \pm 0.9$ & $53.7 \pm 5.9$}
\newcommand{\openpithermalcyclerclose}{$96.3 \pm 0.7$ & $99.7 \pm 0.3$}
\newcommand{\openpithermalcycleropen}{$73.0 \pm 3.1$ & $96.0 \pm 2.1$}
\newcommand{\openpiscrewloose}{$1.3 \pm 1.3$ & $21.3 \pm 1.5$}
\newcommand{\openpipipette}{$4.3 \pm 0.3$ & $42.7 \pm 1.8$}
\newcommand{\openpiinsert}{$2.0 \pm 0.6$ & $40.7 \pm 5.4$}
\newcommand{\openpiscrewtighten}{$0.7 \pm 0.3$ & $2.0 \pm 0.6$}
\newcommand{\openpithermalmixer}{$0.8 \pm 0.2$ & $7.5 \pm 0.6$}
\newcommand{\openpiinsertcentrifuge}{$2.0 \pm 0.6$ & $14.7 \pm 1.3$}
\newcommand{\rdtpickup}{$45.3 \pm 6.8$ & $57.7 \pm 1.2$}
\newcommand{\rdtthermalcyclerclose}{$100.0 \pm 0.0$ & $100.0 \pm 0.0$}
\newcommand{\rdtthermalcycleropen}{$100.0 \pm 0.0$ & $99.0 \pm 0.6$}
\newcommand{\rdtscrewloose}{$2.0 \pm 1.5$ & $2.7 \pm 1.2$}
\newcommand{\rdtpipette}{$0.0 \pm 0.0$ & $0.3 \pm 0.3$}
\newcommand{\rdtinsert}{$0.3 \pm 0.3$ & $2.0 \pm 1.2$}
\newcommand{\rdtscrewtighten}{$3.3 \pm 1.2$ & $8.3 \pm 4.4$}
\newcommand{\rdtthermalmixer}{$0.2 \pm 0.1$ & $1.6 \pm 0.5$}
\newcommand{\rdtinsertcentrifuge}{$1.7 \pm 1.2$ & $1.0 \pm 1.0$}

    \centering
    \caption{VLA evaluation score measured over 100 episodes on AutoBio tasks with increasing level of difficulty. The score is reported in percentage by computing the value and the standard error of the mean over three runs.}
    \label{tab:result}
    \begin{tabular}{
        c
        |
        >{\centering}p{0.12\textwidth}
        >{\centering}p{0.12\textwidth}
        |
        >{\centering}p{0.12\textwidth}
        >{\centering}p{0.12\textwidth}
        |
        >{\centering}p{0.12\textwidth}
        >{\centering\arraybackslash}p{0.12\textwidth}
    }
        \toprule
        \# Demo & 20 & 100 & 20 & 100 & 20 & 100 \\
        \midrule
        \multicolumn{7}{c}{\textbf{Easy level}} \\
        \midrule
        & \multicolumn{2}{c|}{\SB Close thermal cycler lid} & \multicolumn{2}{c|}{\SB Open thermal cycler lid} & \multicolumn{2}{c}{\SB Pick up centrifuge tube} \\
        $\pi_0$ & \openpithermalcyclerclose & \openpithermalcycleropen & \openpipickup \\ 
        RDT & \rdtthermalcyclerclose & \rdtthermalcycleropen & \rdtpickup \\
        \midrule
        \multicolumn{7}{c}{\textbf{Medium level}} \\
        \midrule
        & \multicolumn{2}{c|}{\SB Unscrew centrifuge tube cap} & \multicolumn{2}{c|}{\SB Aspirate with pipette} & \multicolumn{2}{c}{\SB Transfer centrifuge tube} \\
        $\pi_0$ & \openpiscrewloose & \openpipipette & \openpiinsert \\ 
        RDT & \rdtscrewloose & \rdtpipette & \rdtinsert \\
        \midrule
        \multicolumn{7}{c}{\textbf{Hard level}} \\
        \midrule
        & \multicolumn{2}{c|}{\SB Screw on centrifuge tube cap} & \multicolumn{2}{c|}{\SB Operate thermal mixer panel} & \multicolumn{2}{c}{\SB Load centrifuge rotor} \\
        $\pi_0$ & \openpiscrewtighten & \openpithermalmixer & \openpiinsertcentrifuge \\ 
        RDT & \rdtscrewtighten & \rdtthermalmixer & \rdtinsertcentrifuge \\
        \bottomrule
    \end{tabular}
\end{table}

Policies are evaluated over 100 episodes per task, with final scores (mean ± SEM across runs) reported as percentages in Table \ref{tab:result}. Our results reveal distinct characteristics of both VLAs from algorithmic and task-specific perspectives. While neither model demonstrates clear superiority, RDT shows more consistent performance on easy tasks, whereas $\pi_0$ excels in challenging scenarios. This likely stems from $\pi_0$'s PaliGemma-style architecture with joint attention across modalities and fully trainable weights, compared to RDT's static pretrained encoders. This architectural advantage enables $\pi_0$ to better adapt to complex tasks like \textbf{Operate thermal mixer panel}.

Data scaling effects differ markedly: $\pi_0$ benefits from more data across most tasks, while RDT shows limited improvement. This aligns with their architectures - $\pi_0$'s 3B trainable parameters are more data-hungry, whereas RDT's frozen backbones constrain its learning capacity.

Task difficulty rankings are validated by results. Easy task failures primarily involve gripper slippage, while medium and hard tasks reveal precision limitations in current VLAs' imitation learning approach. Compounding errors in precise manipulations (e.g., screwing/unscrewing) highlight the need to explore reinforcement learning or similar approaches for improved closed-loop performance.

Language understanding limitations emerge in \textbf{Transfer centrifuge tube}, where policies frequently select incorrect slots. The challenge intensifies in \textbf{Operate thermal mixer panel}, where models struggle to connect language instructions with visual feedback, exacerbated by low input resolutions that obscure display numbers, leading to more oscillating loss curves (see appendix) than other tasks. This suggests the need for more efficient high-resolution vision processing pipeline.

Visual reasoning demands in \textbf{Aspirate with pipette} (liquid level sensing) and \textbf{Load centrifuge rotor} (symmetry maintenance) expose current VLA limitations. The models fail to adapt to visual cues effectively, and current memoryless architecture faces difficulty when reasoning targets leave the camera view. This indicates the necessity for visual chain-of-thought reasoning and memory mechanisms to maintain execution consistency.

\subsection{Concatenation for Long-Horizon Task}
Human operators naturally decompose complex experiments into atomic subtasks, and execute step by step—a capability we test by combining \textbf{Close/Open thermal cycler lid} episodes (200 in total) to evaluate trajectory concatenation. We assess performance by executing one subtask to completion before switching prompts (Table \ref{tab:long-horizon}).

\begin{table}[h]
    \centering
    \caption{Trajectory concatenation evaluation result}
    \label{tab:long-horizon}
    \begin{tabular}{c|cccc}
        \toprule
        & \textbf{Close} & \textbf{Open} & \textbf{Close-Open} & \textbf{Open-Close} \\
        \midrule
        $\pi_0$ & $100.0 \pm 0.0$ & $96.0 \pm 0.6$ & $3.0 \pm 1.2$ & $87.3 \pm 2.7$ \\
        RDT & $98.7 \pm 0.3$ & $88.0 \pm 3.2$ & $4.3 \pm 2.8$ & $8.7 \pm 2.2$\\
        \bottomrule
    \end{tabular}
\end{table}

After training on mixed data, the performance of RDT drops slightly on the two original tasks \textbf{Close} and \textbf{Open}, while $\pi_0$'s performance mostly remains the same. Regarding concatenated tasks, since the terminal robot pose of \textbf{Open} aligns well with \textbf{Close}'s trajectory, $\pi_0$ could achieve reasonable transition in \textbf{Open-Close}, yet RDT fails to effectively interpolate in-between. \textbf{Close-Open} fails more frequently for both models, due to handle grip orientation differences in demonstration at transition location. This suggests current VLAs primarily memorize trajectories during fine-tuning, with limited ability to generalize or smoothly transition between related tasks.

\section{Limitations and Future Work}
Our work has several limitations that highlight directions for future research.

\paragraph{Simulation Fidelity and Efficiency.} The reliance on Blender Cycles for photorealistic rendering introduces a computational bottleneck, with rendering times $\sim$100× slower than MuJoCo renderer. This is mitigated by the low-resolution inputs of current VLA models, enabling feasible data collection (e.g., 100k frames in hours). However, as tasks like \textbf{Operate thermal mixer panel} reveal, high-resolution perception may be critical for fine-grained manipulation. Future work will explore optimizations such as hybrid EEVEE-Cycles pipelines to balance speed and visual fidelity.

\paragraph{Task and Asset Scalability.} While AutoBio includes biologically grounded tasks, the manual effort required for asset digitization and protocol design limits the diversity of instruments and workflows. We plan to (1) standardize digitization workflow and task templates, (2) employ generative methods for automated asset and task synthesis, and (3) compose long-horizon tasks from modular primitives to support complex, reusable experimental workflows.

\paragraph{Real-World Transfer.} AutoBio’s physics models and rendering, though carefully designed, may not fully capture the stochasticity and perceptual noise of real-world labs. Bridging this sim-to-real gap will require improvements in material modeling, actuator dynamics, and domain randomization.

\section{Summary}


This paper introduce AutoBio, a simulation framework and benchmark designed to evaluate robotic automation in biology laboratory environments. By extending laboratory asset modeling, physics simulation, and rendering capabilities, the AutoBio simulator streamlines the biological experiment simulation inspired by practical operation primitives. Built upon this simulator, we propose the AutoBio benchmark with manipulation tasks across three levels of difficulty, to systematically assess the capabilities of vision-language-action (VLA) models. The experimental results show critical limitations of current VLA models in precision manipulation, instruction following and visual reasoning, and suggest potential improvements in model architecture and training methodologies.

\bibliographystyle{plain}
\bibliography{refs}

\appendix
\section{AutoBio Simulation}
\subsection{Simulation-ready assets}
\begin{figure}[h]
    \centering
    \includegraphics[width=0.88\linewidth]{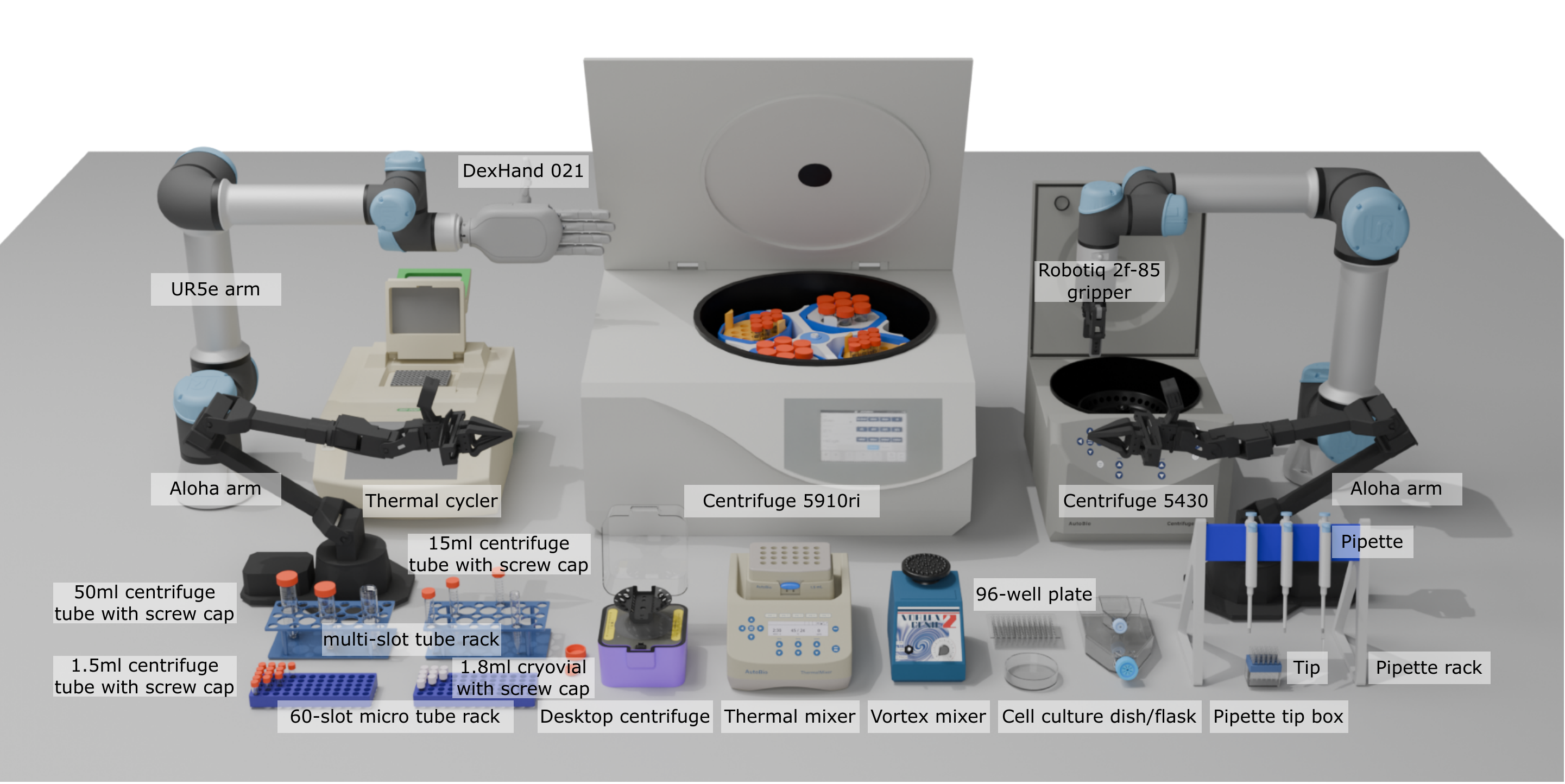}
    \caption{The overview of AutoBio's simulation-ready digital assets.}
    \label{fig:family}
\end{figure}

\subsection{Thread Mechanism}
A parameterized circular helix in three-dimensional space can be mathematically expressed as:
\begin{equation*}
    H(t; r_1, p)=[x,y,z]^\top=[r_1\cos t, r_1\sin t, pt]^\top.
\end{equation*}
In practical applications, a helix is always bounded. For the above helix formulation, we denote these bounds $2\pi l \leq t \leq 2\pi h$, where $l$ and $h$ represent the starting and ending thread counts measured from the zero position. Within this framework, the approximate SDF becomes:
\begin{gather*}
    t_{0}=\operatorname{atan2}(P_y, P_x), \\
    k=\left\lfloor{\frac{P_z-t_{0}p}{2\pi{}p}}\right\rceil, \quad
    l'=\left\lceil \frac{l - t_0}{2\pi}\right\rceil, \quad
    h'=\left\lfloor \frac{h - t_0}{2\pi}\right\rfloor, \\
    SDF(P) = \begin{cases}
        d_P(2k\pi+t_0) & l' \leq k \leq h' \\
        \min \{ d_P(2\pi l), d_P(2l'\pi+t_0) \} & k < l' \\
        \min \{ d_P(2\pi h), d_P(2h'\pi+t_0) \} & h' < k \\
    \end{cases},
\end{gather*}
where $d_P(t)$ represents the Euclidean distance between point $P$ and the corresponding position on the helix parameterized by $t$:
\begin{equation*}
    d_P(t) = \left\|H\left(t;r_1, p\right)-P\right\|_2.
\end{equation*}
This approximation remains valid when the helix angle is sufficiently small.

In practical implementation, the plugin has been adapted to MuJoCo as an \verb|sdf| extension. While alternative methods exist to approximate screw motion—such as parenting the nut to the bolt within a kinematic tree and adding a coupled hinge joint and slide joint constrained by screw motion—these approaches present two significant limitations: (1) The self-locking mechanism does not naturally emerge from such couplings and proves challenging to implement artificially; (2) Modeling transitional states between fully coupled and completely free bolt-nut pairs becomes problematic. By employing SDF to characterize these interactions in a manner more faithful to real-world physics, we address both challenges gracefully.

\subsection{Detent Mechanism}
The detent mechanism is modeled using a passive force function $f(q, \dot{q})$that depends exclusively on the generalized position and velocity—states intrinsic to any dynamic system. For a knob with $n$ discrete positions, the force calculation follows:
\begin{equation*}
    f(q, \dot{q}) = -k(q - q_j) - \lambda \dot{q},
\end{equation*}
where the target position index is determined by:
\begin{equation*}
    j = \operatorname*{argmin}_i{|q - q_i|}, \quad i = 0, \ldots, n-1.
\end{equation*}
In practical implementation, this mechanism is adapted to MuJoCo as a \verb|passive force| extension.

\subsection{Eccentric Mchanism}
Planar eccentric motion can be mathematically represented through a planar transformation matrix, which admits the following decomposition:
\begin{equation}
    \begin{pmatrix}
        1 & 0 & t\cos\theta\\
        0 & 1 & t\sin\theta\\
        0 & 0 & 1
    \end{pmatrix}=
    \begin{pmatrix}
        \cos\theta & -\sin\theta & t\cos\theta\\
        \sin\theta & \cos\theta  & t\sin\theta\\
        0          & 0           & 1
    \end{pmatrix}
    \begin{pmatrix}
        \cos\theta  & \sin\theta  & 0\\
        -\sin\theta & \cos\theta  & 0\\
        0           & 0           & 1
    \end{pmatrix},
\end{equation}
where $t$ denotes the throw of the eccentric. The right-hand side matrices correspond to the motions of two negatively-coupled hinge joints. Instruments employing eccentric mechanisms, such as vortex mixers, typically operate at high RPMs. Under these conditions, coupling two joints through equality constraints proves more robust than alternative solutions.

\subsection{Quasi-static Liquid}
The surface normal dynamics are governed by a damped spherical pendulum system. Following analytical mechanics conventions, we present the system's Lagrangian and generalized force as:
\begin{align*}
    L&=\frac{1}{2} m l^{2}\left(\dot{\phi}^{2} \sin^{2}{\theta} + \dot{\theta}^{2}\right) + m l\left(g_{x} \sin{\theta} \cos{\phi} + g_{y} \sin{\phi} \sin{\theta} - g_{z} \cos{\theta}\right), \\
    Q&=[-\lambda_\phi \dot{\phi}, -\lambda_\theta \dot{\theta}]^\top.
\end{align*}
$[\phi, \theta]$ are the spherical coordinates of the normal vector and the generalized coordinates of the system. $l$ represents the characteristic length, and $m$ is system mass (which ultimately cancels out in the final equations). The vector $g=[g_x,g_y,g_z]^\top$ captures time-varying accelerations resulting from both gravity and inertial forces. $\lambda_\phi,\lambda_\theta$ are configurable damping coefficients. Applying the Euler-Lagrange equation:
\begin{equation*}
    \frac{\dif}{\dif t}\frac{\dif{}L}{\dif{}\dot{q_i}} - \frac{\dif{}L}{\dif{}q_i} = Q_i,
\end{equation*}
we derive the following system of ordinary differential equations:
\begin{align*}
    \frac{\dif{}}{\dif{}t} \phi &= \dot\phi, \\
    \frac{\dif{}}{\dif{}t} \theta &= \dot\theta, \\
    \frac{\dif{}}{\dif{}t} \dot\phi &= \frac{- 2ml^2 v_{\phi} v_{\theta} \sin\theta\cos\theta + ml \left(- g_{x} \sin\phi + g_{y} \cos\phi\right) \sin\theta - \lambda_{\phi} v_{\phi}}{ml^{2} \sin^{2}\theta}, \\
    \frac{\dif{}}{\dif{}t} \dot\theta &= \frac{ml^{2} v_{\phi}^{2} \sin\theta\cos\theta + ml \left(g_{x} \cos\phi \cos\theta + g_{y} \sin\phi \cos\theta + g_{z} \sin\theta\right) - \lambda_{\theta} v_{\theta}}{ml^2}.
\end{align*}
In practical implementation, to avoid numerical instabilities when the system approaches simple pendulum behavior, we modify the denominator in the $\frac{\dif{}}{\dif{}t} \dot\phi$ equation to $ml^2 \max\{\sin^{2}\theta, \epsilon\}$. The system initializes with states aligned to gravity direction and zero velocity. The normal direction components are computed as:
\begin{align*}
    x &= \sin\theta \cos\phi, \\
    y &= \sin\theta \sin\phi, \\
    z &= -\cos\phi.
\end{align*}
Following surface normal determination at each timestep, we compute the liquid body as the intersection between the oriented halfspace and the container's inner surface. This implementation involves computing triangle-plane relations for the manifold mesh. Surface height computation proceeds through volume conservation constraints, employing the previous height as an initial guess before refining via Newton-Bisect search.

\subsection{Rendering}
The MuJoCo renderer utilizes the legacy OpenGL fixed-function pipeline for scene rendering, which offers limited customization capabilities. Notably, its reliance on fixed-function lighting creates visual inconsistencies between textured and untextured surfaces under specular and ambient lighting conditions. For this renderer, we provide best-effort support, with certain scene elements (such as liquid representations) appearing in simplified forms to minimize visual artifacts.

We develop a bridging mechanism to translate MuJoCo's scene definition (\verb|MjModel|) and simulation state (\verb|MjData|) into Blender's environment. While conceptually similar to MuJoCo's ongoing USD (Universal Scene Description) exporter, our solution specifically accommodates our workflow by enabling: (1) Importation of tuned assets from Blender gallery files, and (2) Direct generation of fully configured Blender scenes capable of rendering features beyond MuJoCo's representational capacity.

\subsection{Reactive UI}
Our user interfaces currently operate in retained mode, meaning it only repaints when changes occur. For the MuJoCo renderer, UI updates in the 3D environment are achieved through calls to \verb|mjr_uploadTexture| following repainting. In the Blender renderer, the UI is passed to the \verb|Image Texture| node as either static images or video streams.

\section{Experiment Details}
\subsection{Demonstration Synthesis}
For motion planning, AutoBio integrates inverse kinematics (IK) solvers with time-optimal path parameterization (TOPP) to produce collision-free trajectories that respect dynamic constraints. Robot control for these demonstrations is achieved via proportional-derivative (PD) controllers with tunable gains, balancing precision and stability.

For the Aloha arm, since it forms a spherical wrist, we implement an analytical IK solver to improve robustness. For UR5e arm, we use our optimization-based IK solver due to its larger joint range.

\subsection{Benchmark Tasks}
In the experiment section, we evaluated VLA capabilities across 9 AutoBio tasks. Below we describe the details of each task.

\begin{figure}[H]
    \includegraphics[width=\linewidth]{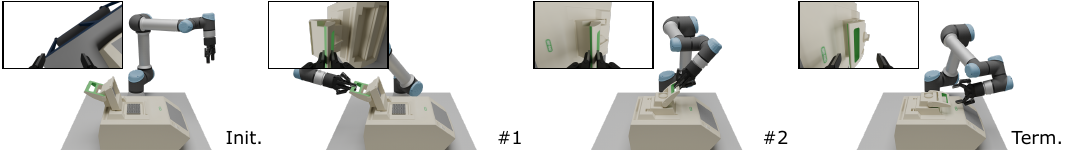}
    \begin{tcolorbox}[width=\linewidth, sharp corners=all, colback=white!95!black, before skip=2px, left=4pt,right=4pt,top=2pt,bottom=2pt, boxrule=0pt, colframe=white]%
    \textbf{Close thermal cycler lid} (easy, $\sim$\qty{20}{s}): \emph{``close the lid of the thermal cycler''} Close and lock the thermal cycler lid using a \textit{UR5e-Robotiq} robot, testing trajectory following for articulated object manipulation. The joint angles of the robotic arms are perturbed to add randomness to the task. This randomization also applies to all tasks below.
    \end{tcolorbox}
\end{figure}

\begin{figure}[H]
    \includegraphics[width=\linewidth]{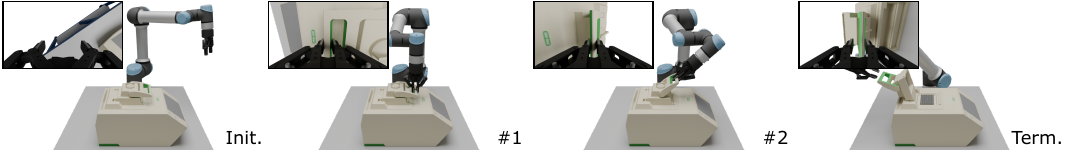}
    \begin{tcolorbox}[width=\linewidth, sharp corners=all, colback=white!95!black, before skip=2px, left=4pt,right=4pt,top=2pt,bottom=2pt, boxrule=0pt, colframe=white]
    \textbf{Open thermal cycler lid} (easy, $\sim$\qty{17}{s}): \emph{``open the lid of the thermal cycler''} Unlock and open the thermal cycler lid using a \textit{UR5e-Robotiq} robot, testing trajectory following for articulated object manipulation.
    \end{tcolorbox}
\end{figure}

\begin{figure}[H]
    \includegraphics[width=\linewidth]{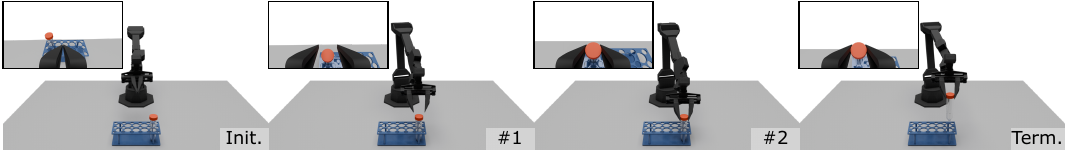}
    \begin{tcolorbox}[width=\linewidth, sharp corners=all, colback=white!95!black, before skip=2px, left=4pt,right=4pt,top=2pt,bottom=2pt, boxrule=0pt, colframe=white]
    \textbf{Pick up centrifuge tube} (easy, $\sim$\qty{10}{s}): \emph{``pick up the centrifuge tube on the rack''} Pick up a centrifuge tube from its rack using an \textit{Aloha} robot, focusing on basic visual positioning. The tube is placed in a random rack slot to promote generalization. This also applies to all tasks below involving tubes and racks.
    \end{tcolorbox}
\end{figure}

\begin{figure}[H]
    \includegraphics[width=\linewidth]{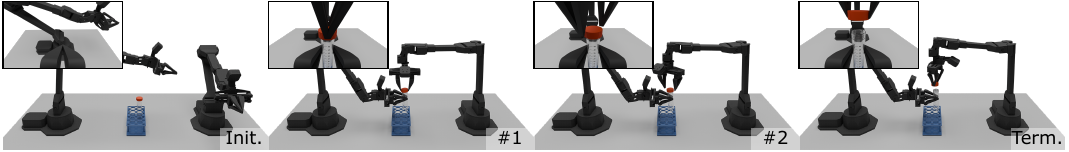}
    \begin{tcolorbox}[width=\linewidth, sharp corners=all, colback=white!95!black, before skip=2px, left=4pt,right=4pt,top=2pt,bottom=2pt, boxrule=0pt, colframe=white]
    \textbf{Unscrew centrifuge tube cap} (medium, $\sim$\qty{27}{s}): \emph{``dual-Aloha arms unscrewing centrifuge tube cap: one stabilizes tube while the other twists cap''} Remove the centrifuge tube cap using two \textit{Aloha} robots. This evaluates dual-arm coordination and manipulation precision.
    \end{tcolorbox}
\end{figure}

\begin{figure}[H]
    \includegraphics[width=\linewidth]{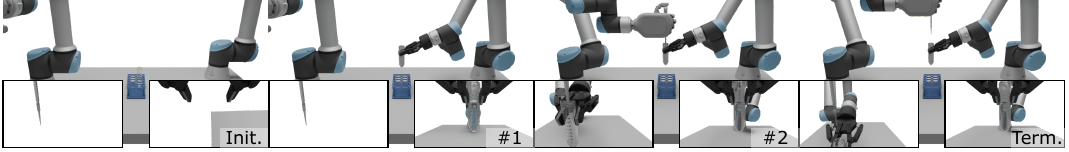}
    \begin{tcolorbox}[width=\linewidth, sharp corners=all, colback=white!95!black, before skip=2px, left=4pt,right=4pt,top=2pt,bottom=2pt, boxrule=0pt, colframe=white]
    \textbf{Aspirate with pipette} (medium, $\sim$\qty{14}{s}): \emph{``dual-UR5e pipetting: one arm lifts centrifuge tube, the other aligns pipette tip and aspirates liquid''} Aspirate liquid from a tube using a \textit{UR5e-Robotiq} (holder) and \textit{UR5e-DexHand} (pipette operator). This tests dual-arm coordination, precision, and visual reasoning. The liquid volume in the tube is randomized to test liquid level sensing capabilities.
    \end{tcolorbox}
\end{figure}

\begin{figure}[H]
    \includegraphics[width=\linewidth]{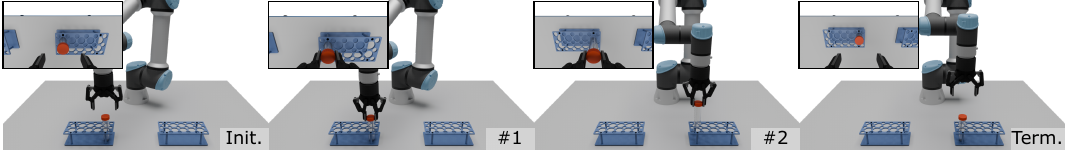}
    \begin{tcolorbox}[width=\linewidth, sharp corners=all, colback=white!95!black, before skip=2px, left=4pt,right=4pt,top=2pt,bottom=2pt, boxrule=0pt, colframe=white]
    \textbf{Transfer centrifuge tube} (medium, $\sim$\qty{11}{s}): \emph{``pick up the centrifuge tube and move it to the other rack, row \texttt{\{target\_row\}}, column \texttt{\{target\_col\}}''} Transfer a tube to a specified rack slot using a \textit{UR5e-Robotiq} robot, requiring precise manipulation, visual positioning, and language instruction following. The target rack slot is randomized to verify instruction following.
    \end{tcolorbox}
\end{figure}

\begin{figure}[H]
    \includegraphics[width=\linewidth]{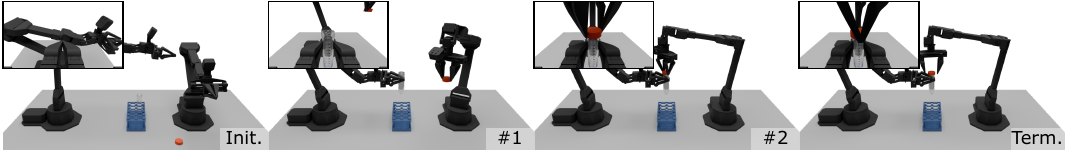}
    \begin{tcolorbox}[width=\linewidth, sharp corners=all, colback=white!95!black, before skip=2px, left=4pt,right=4pt,top=2pt,bottom=2pt, boxrule=0pt, colframe=white]
    \textbf{Screw on centrifuge tube cap} (hard, $\sim$\qty{31}{s}): \emph{``dual-Aloha arms screwing on centrifuge tube cap: one grips tube while the other twists cap''} The inverse of unscrewing, but with stricter precision requirements for proper alignment.
    \end{tcolorbox}
\end{figure}

\begin{figure}[H]
    \includegraphics[width=\linewidth]{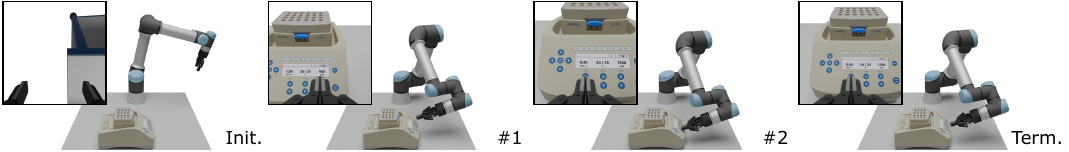}
    \begin{tcolorbox}[width=\linewidth, sharp corners=all, colback=white!95!black, before skip=2px, left=4pt,right=4pt,top=2pt,bottom=2pt, boxrule=0pt, colframe=white]
    \textbf{Operate thermal mixer panel} (hard, $\sim$\qty{16}{s}): \emph{``Adjust thermal mixer parameters, with speed set to \texttt{\{set\_rpm\}} rpm, temperature set to \texttt{\{set\_temp\}} \textdegree{}C, and time set to \texttt{\{set\_time\}} seconds''} Set parameters (time, temperature, frequency) on a mixer panel using a \textit{UR5e-Robotiq} robot following language instruction and UI feedback, evaluating visual reasoning, language understanding, and high-precision manipulation. The parameters are randomized to test cross-modal reasoning.
    \end{tcolorbox}
\end{figure}

\begin{figure}[H]
    \includegraphics[width=\linewidth]{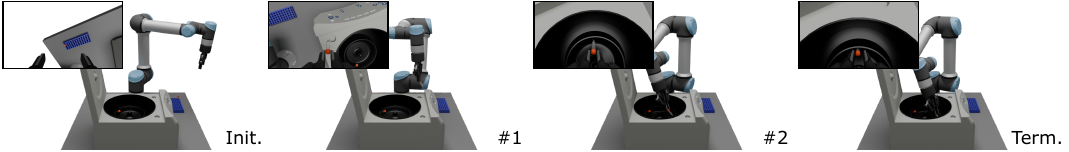}
    \begin{tcolorbox}[width=\linewidth, sharp corners=all, colback=white!95!black, before skip=2px, left=4pt,right=4pt,top=2pt,bottom=2pt, boxrule=0pt, colframe=white]
    \textbf{Load centrifuge rotor} (hard, $\sim$\qty{11}{s}): \emph{``Insert a second centrifuge tube into the slot that is symmetrically opposite to the currently placed tube''} Load a tube into the correct rotor slot while maintaining symmetry with the existing tube, testing advanced visual reasoning and precise positioning. The target rotor slot are randomized by setting the currently placed tube to different slots, and the rotor angle are also randomized.
    \end{tcolorbox}
\end{figure}

The evaluation time limits extended by approximately \qty{50}{\%} compared to demonstration horizon to accommodate imperfect policy execution. All tasks except \textbf{Operate thermal mixer panel} are scored binarily (1 for success, 0 for exceeding time limit) based on pose and contact requirements at checkpoints and terminal states. For \textbf{Operate thermal mixer panel}, we implement a weighted relative progress score to better reflect performance difference due to observed policy difficulties:
\begin{equation*}
    \text{Score} = \sum_{i=1}^{3}{w_i \max\left\{1 - \frac{|\text{final}_i - \text{target}_i|}{|\text{initial}_i - \text{target}_i|}, 0\right\}},
\end{equation*}
normalized to $[0,1]$.

\subsection{Baselines}
We summarize key differences of our baselines ($\pi_0$ and RDT) in Table~\ref{tab:vla}, where some of them (proprioception dimensions, image history, normalization) lead to difference in data adaptation.

\begin{table}[h]
\centering
\caption{Key differences between $\pi_0$ and RDT baselines}
\label{tab:vla}
\begin{tabular}{lll}
\toprule
                  & $\pi_0$              & RDT                     \\
\midrule
Architecture      & Decoder-only         & Encoder-decoder         \\
Vision backbone   & SigLIP So400m/14 224 & SigLIP So400m/14 384    \\
Language backbone & Gemma 2B+300M        & T5 v1.1 XXL             \\
Trainable weights & Full                 & V \& L backbone frozen  \\
Action sampling   & Flow matching        & Diffusion               \\
Normalization     & All                  & Gripper width only      \\
State dimension   & 32                   & 128                     \\
Action horizon    & 50                   & 64                      \\
Image history     & 0 (None)             & 1                       \\
\bottomrule
\end{tabular}
\end{table}

\subsection{Training curves}
\begin{figure}
    \centering
    \includegraphics[width=\linewidth]{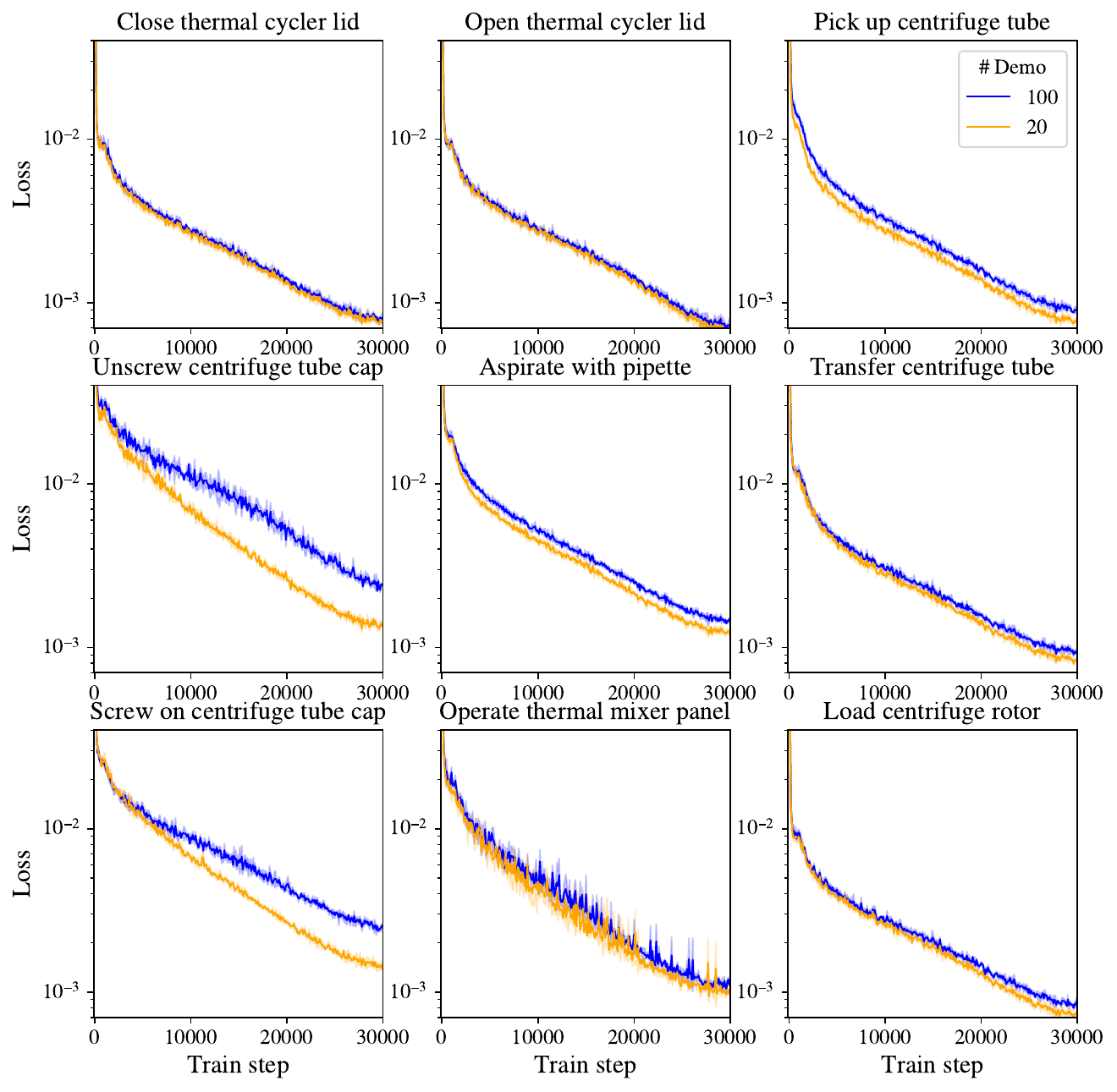}
    \caption{Loss curves of $\pi_0$ on 9 tasks evaluated. The solid line shows the mean over 3 runs; the shaded region indicates the $\pm1\sigma$ confidence interval.}
    \label{fig:train-pi0}
\end{figure}

\begin{figure}
    \centering
    \includegraphics[width=\linewidth]{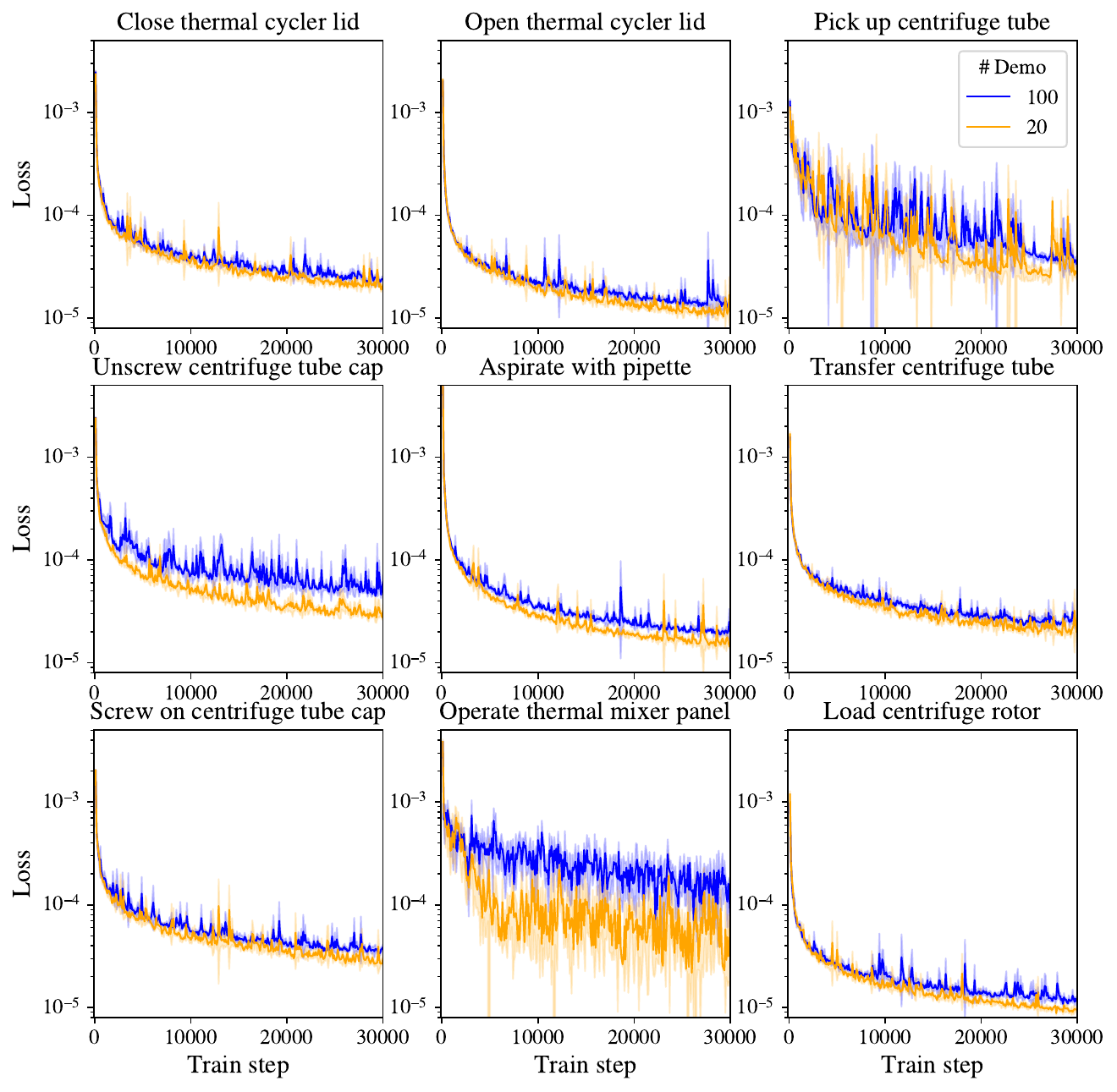}
    \caption{Loss curves of RDT on 9 tasks evaluated. The solid line shows the mean over 3 runs; the shaded region indicates the $\pm1\sigma$ confidence interval.}
    \label{fig:train-rdt}
\end{figure}

\begin{figure}
    \centering
    \subfloat[$\pi_0$]{\includegraphics[width=0.45\linewidth]{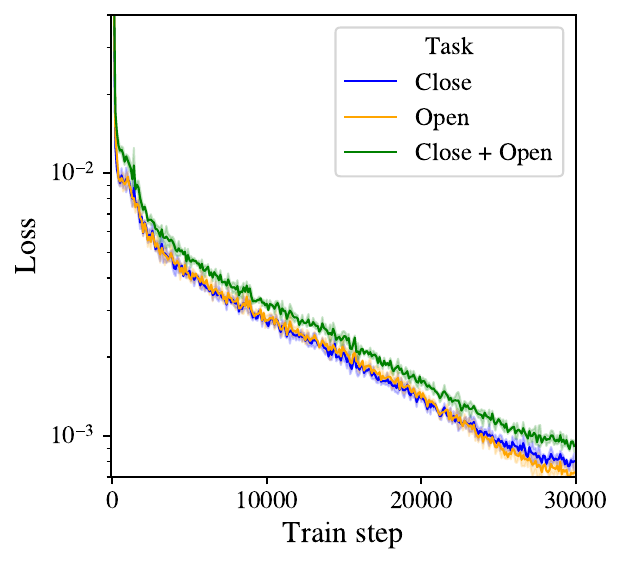}}\hfill
    \subfloat[RDT]{\includegraphics[width=0.45\linewidth]{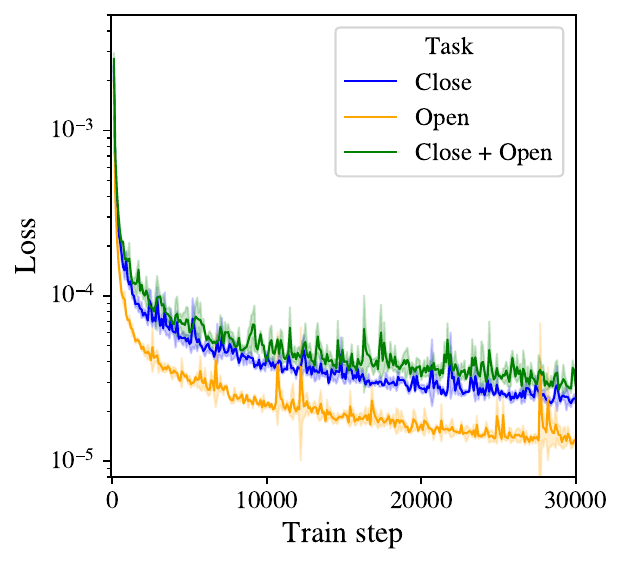}}
    \caption{Loss curves for the task concatenation experiment. The solid line shows the mean over 3 runs; the shaded region indicates the $\pm1\sigma$ confidence interval.}
    \label{fig:train-concat}
\end{figure}

Figures~\ref{fig:train-pi0}, \ref{fig:train-rdt}, and \ref{fig:train-concat} present the training loss curves for $\pi_0$, RDT, and the task concatenation experiment, respectively. Several observations can be made:
\begin{itemize}[leftmargin=*]
\setlength\itemsep{0em}
    \item The training loss curves of $\pi_0$ generally exhibit smoother convergence compared to those of RDT.
    \item The \textbf{Operate thermal mixer panel} task induces noticeable oscillations in the training curves for both $\pi_0$ and RDT.
    \item For $\pi_0$, dual-arm tasks—particularly those involving the two screw operations—tend to overfit more easily when trained on limited data.
\end{itemize}

\end{document}